\documentclass{article}
\usepackage{tikz}    
\usepackage[final,nonatbib]{nips_2016}
\usepackage{enumitem}
\usepackage[utf8]{inputenc} 
\usepackage[T1]{fontenc}    
\usepackage{hyperref}       
\usepackage{url}            
\usepackage{booktabs}       
\usepackage{amsmath}       
\usepackage{caption}       
\usepackage{subcaption}       
\usepackage{amsfonts}       
\usepackage{nicefrac}       
\usepackage{microtype}      

\usepackage[backend=bibtex]{biblatex}
\bibliography{smallbiblio}
\tikzset{>=latex}

\title{Learning to learn with backpropagation of Hebbian plasticity}

\author{
    Thomas Miconi\\
    The Neurosciences Institute\\
    La Jolla, CA, USA\\
    \texttt{miconi@nsi.edu}
}

\begin{document}

\maketitle

\begin{abstract}

Hebbian plasticity is a powerful principle that allows biological brains to
learn from their lifetime experience.  By contrast, artificial neural networks
trained with backpropagation generally have fixed connection weights that do
not change once training is complete.  While recent methods can endow neural
networks with long-term memories, Hebbian plasticity is currently not amenable
to gradient descent. Here we derive analytical expressions for activity
gradients in neural networks with Hebbian plastic connections. Using these
expressions, we can use backpropagation to train not just the baseline weights
of the connections, but also their plasticity. As a result, the networks
``learn how to learn''  in order to solve the problem at hand: the trained
networks automatically perform fast learning of unpredictable environmental
features during their lifetime, expanding the range of solvable problems. We
test the algorithm on various on-line learning tasks, including pattern completion, one-shot
learning, and reversal learning.  The algorithm successfully learns how to
learn the relevant associations from one-shot instruction, and fine-tunes the
temporal dynamics of plasticity to allow for continual learning in response to
changing environmental parameters. We conclude that backpropagation of Hebbian
plasticity offers a powerful model for lifelong learning.

\end{abstract}

\section{Introduction}

Living organisms endowed with neural systems exhibit remarkably complex behaviors.
While much of this complexity results from evolutionary learning over millions
of years, it also results from the ability of neural systems to learn from
experience during their lifetime. Indeed, this ability for lifelong learning is
itself a product of evolution, which has fashioned not just the overall
connectivity of the brain, but also the plasticity of these connections. 

Lifetime learning is beneficial for several reasons. For one thing,
many environmental features can simply not be predicted at birth, and/or change
over time (e.g. the
position of food sources, the identifying features of specific individuals, etc.), 
requiring learning from experience in contact with the
environment. Furthermore, even for predictable environmental features, much of
the information necessary to produce adaptive behavior can be obtained ``for
free'' by learning from the environment, thus removing a potentially huge chunk
of the search space that evolution must explore. For example, the connectivity
of primary visual cortex is fashioned by Hebbian plasticity rather than having
each single connection genetically specified \cite{espinosa2012development}, allowing a huge number of cells
to organize into a powerful, reliable information-processing system with
minimal genetic specification.

Lifetime long-term plasticity in living brains generally follows the Hebbian principle: a
cell that consistently contributes in making another cell fire will build a
stronger connection to that cell. Note that this generic principle can be
implemented in many different ways, including covariance learning, instar and
outstar rules, BCM learning, etc. (see \cite{Vasilkoski2011-ww} and references therein). 

Backpropagation can train neural networks to perform remarkably complex
tasks. However, it is generally used to train fixed-weights networks, with no further changes in connectivity after training. 
Several
methods have been proposed to make lifelong learning amenable to
backpropagation, including most recently neural Turing machines
\cite{Graves2014-ch,Santoro2016-jn} and
memory networks \cite{Sukhbaatar2015-ly}. However, it would be useful to incorporate the powerful, well-studied
principle of Hebbian plasticity in backpropagation
training.

Here we derive analytical expressions for activity gradients in neural networks
with Hebbian plastic connections. Using these expressions, we can use
backpropagation to train not just the baseline weights of the connections, but
also their plasticity. This allows backpropagation to ``learn how to learn'',
in order to solve general types of problems with unpredictable features, rather
than specific instances.

All software used for the present paper is available at
\texttt{http://github.com/thomasmiconi}.

\section{Networks with Hebbian synapses}

We consider networks where the strength of each connection can vary according to
Hebbian plasticity over the course of the network's lifetime. We will arrange things so that each network is fully specified
by fixed parameters which determine both the baseline weight \emph{and} the degree of plasticity of each connection.
After training, these parameters are fixed and unchanging over the network's
lifetime, but govern the way in which
each connection changes over time, as a result of experience, according to Hebbian plasticity. 

To model Hebbian plasticity, we maintain a time-dependent quantity for each
connection in the network, which we call the \emph{Hebbian trace} for this
connection.  As noted above, there are many possible expressions for Hebbian
plasticity \cite{Vasilkoski2011-ww}.  In this paper, we use the simplest stable form of Hebbian trace,
namely, the running average of the product of pre- and post-synaptic
activities.  Thus, for a given target cell, the Hebbian trace associated with
its $k$-th incoming connection is defined as follows:

\begin{equation}
\label{eq:hebb}
Hebb_k(t) = (1-\gamma) * Hebb_k(t-1) + \gamma * x_k(t) * y(t)
\end{equation}

where $y(t)$ is the activity of the post-synaptic cell, $x_k(t)$ is the activity
of the pre-synaptic cell, and $\gamma$ is a time constant. While other
expressions of Hebbian plasticity are possible, this simple form turns out to be
adequate for our present purposes and simplifies the mathematics.

The Hebbian trace is maintained automatically, independently of network
parameters, for each connection. Given this Hebbian trace, the
actual strength of the connection at time $t$ is determined by two fixed parameters: a fixed
weight $w_k$, which determines the ``baseline'', unchanging component of the
connection; and a \emph{plasticity parameter} $\alpha_k$, which specifies how
much the Hebbian trace influences the actual connection. More formally, the
response $y$ of a given cell can be written as a function of its inputs as
follows:

\begin{equation}
\label{eq:y}
y(t) = \tanh\left\{\sum_{k \in inputs}[w_k x_k(t) + \alpha_k Hebb_k(t) x_k(t)] +
b\right\}
\end{equation}

where $b$ is a bias parameter.

\section{Gradients}

In order to use backpropagation, we must find the gradient of $y$ over the
$w_k$ and $\alpha_k$ parameters. Importantly, these gradients will necessarily
involve activities at previous times, since under plasticity activity at time
$t$ influences activity at future times $t+n$ due to its effects on Hebbian
traces. Fortunately, these gradients turn out to have a simple, recursive form. 

Temporarily omitting the $\tanh$ nonlinearity (see below), we get the
following expressions: 

\begin{equation}
\label{eq:gradw}
\frac{\partial y(t_z)}{\partial w_k} = x_k(t_z) + \sum_{l \in inputs}[\alpha_l
x_l(t_z) \sum_{t_u<t_z}(1-\gamma) \gamma^{t_z-t_u} x_l(t_u) \frac{\partial
y(t_u)}{\partial w_k}]
\end{equation}

\begin{equation}
\label{eq:gradalpha}
\frac{\partial y(t_z)}{\partial \alpha_k} = x_k(t_z) Hebb_k(t_z) + \sum_{l \in inputs}
[\alpha_l x_l(t_z) \sum_{t_u<t_z}(1-\gamma) \gamma^{t_z-t_u} x_l(t_u) \frac{\partial
y(t_u)}{\partial \alpha_k}]
\end{equation}

(See Appendix for a full derivation.)

These equations express the gradient of $y(t_z)$ as a function of the gradients
of $y(t_z<t_u)$, that is, recursively.

In each of these equations, the summand over previous times $t_u<t_z$ is
essentially the partial derivative of the Hebbian traces at time $t_{z}$
with respect to either $w_k$ (Eq. \ref{eq:gradw}) or $\alpha_k$ (Eq.
\ref{eq:gradalpha}). Since the Hebbian trace is the exponential average of
previous products of $x$ and $y$, these partial derivatives turn out to be
sums of the previous gradients of $y$ over the corresponding parameter,
multiplied by the concomitant activity of the input cell $x_k$ (the $\gamma$ terms
account for the exponential decay of the running average). Thus, the gradient at
time $t_z$ is a function of (the weighted sum of) the gradients at times
$t_u<t_z$.

Note that the sum is over the Hebbian traces of \emph{all} inputs to $y$, not just the
one associated with connection $k$ for which we are computing the gradient. This is because the values of $w_k$ and
$\alpha_k$, by affecting $y$, also influence the Hebbian traces of all other
connections to $y$ - which will in turn further affect $y$ at later times. This effect must be accounted for in the above gradients.

The above expression omits the $\tanh$ nonlinearity: it really provides the
gradient of the expression within the curly braces in Eq. \ref{eq:y}, that is,
the ``raw'' output (call it $y_{raw}$) provided by incoming excitation and biases. To obtain
the full gradient $\frac{\partial y(t_z)}{\partial w_k}$, we simply rewrite $y$ as $y =
\tanh(y_{raw})$ and apply the chain rule: $\frac{\partial y}{\partial w_k} =
\frac{\partial \tanh(y_{raw})}{\partial y_{raw}} \frac{\partial y_{raw}}{\partial
w_k} = (1 - y^2)\frac{\partial y_{raw}}{\partial w_k}$, where $\frac{\partial
y_{raw}}{\partial w_k}$ is provided by Eq. \ref{eq:gradw} above (and similarly
for $\frac{\partial
y_{raw})}{\partial \alpha_k}$ 

\section{Experiments}

\subsection{Applying BOHP}

In all tasks described below, lifetime experience is divided into
\emph{episodes}, each of which lasts for a certain number of timesteps. At the
beginning of each episode, all Hebbian traces are initialized to 0. Then,
at each timestep, the network processes an input pattern and produces an output
according to its current parameters, and the Hebbian traces are updated
according to Equation \ref{eq:hebb}. Furthermore, errors and gradients are also
computed. At the end of each episode, the errors and gradients at each timestep
are used to update network parameters (weights and plasticity coefficients)
according to error backpropagation and gradient descent. The whole process
iterates for a fixed number of episodes. For the last 500 episodes, training
stops and networks parameters are frozen.

All source code for these experiments is available at \texttt{http://github.com/thomasmiconi}.

\subsection{Pattern completion}

\begin{figure}
\centering
\begin{subfigure}[t]{0.35\textwidth}
\centering
\includegraphics[scale=0.5]{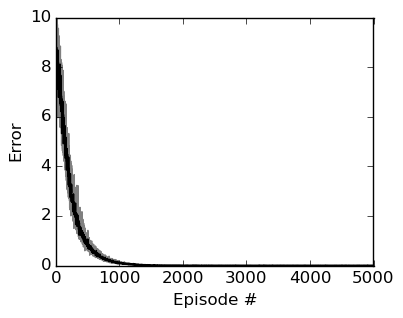}
\subcaption{Training error}
\end{subfigure}
\quad
\begin{subfigure}[t]{0.25\textwidth}
\centering
\begin{tikzpicture}
\begin{scope}[every node/.style={circle,thick,draw}]
\node (O1) at (0,4) {O1};
\node (O2) at (1,4) {O2};
\node (O3) at (2,4) {O3};
\node (I1) at (0,2) {I1};
\node (I2) at (1,2) {I2};
\node (I3) at (2,2) {I3};
\end{scope}
\begin{scope}[
        every node/.style={fill=white,circle},
    every edge/.style={draw=black,very thick}]

    \path [->] (I1) edge   (O1);
    \path [dashed][->] (I1) edge  (O2);
    \path [dashed][->] (I1) edge  (O3);

    \path [->] (I2) edge   (O2);
    \path [dashed][->] (I2) edge  (O1);
    \path [dashed][->] (I2) edge  (O3);
    
    \path [->] (I3) edge   (O3);
    \path [dashed][->] (I3) edge  (O1);
    \path [dashed][->] (I3) edge  (O3);
\end{scope}
\end{tikzpicture}
\subcaption{Typical trained network}
\end{subfigure}
\begin{minipage}[b]{0.25\textwidth}
\centering
\noindent
\begin{itemize}[leftmargin=*] 
\item[]\tikz{\path[very thick,->,draw=black] 
        (0,0) -- (1,0) ;}Excitatory fixed-weight connection
\item[]\tikz{\path[very thick,->,dashed,draw=black] 
        (0,0) -- (1,0) ;}Excitatory plastic connection
\end{itemize}
\end{minipage}
\caption{Results for the pattern completion experiment. (a) Mean absolute error per timestep over each episode, for mutually
exclusive stimuli. The dark line indicates median error
over 20 runs, while shaded areas indicate interquartile range. For the last 500
episodes, training is halted and parameters are frozen. (b) Schema of a
typical network after training. Only 3 elements shown for clarity (actual
pattern size: 8 elements). }
\label{fig:completion}
\end{figure}
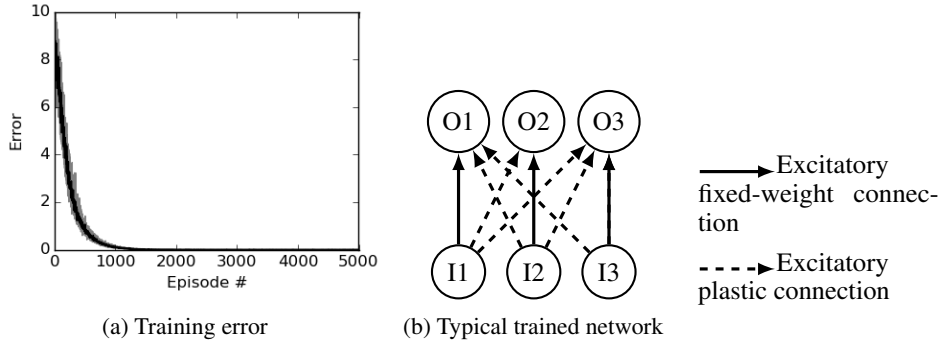

To test the BOHP method, we first apply it to a task for which Hebbian learning
is known to be efficient, namely, pattern completion. The network is composed
of an input and an output layer, each having $N$ neurons.  In every episode,
the network is first exposed to a random binary vector of length $N$ with at
least one nonzero element. This binary vector represents the pattern to be
learned.  Then, at the next timestep, a partial pattern containing only one of
the non-zero bits of the pattern (all other bits set to 0) is presented. The
task of the network is to produce the full pattern in the output layer. The
error for each episode is the Manhattan distance between the network's output
at the second time step and the full pattern (network response during the first
step is ignored).

The algorithm quickly and reliably learns to solve the task (Figure
\ref{fig:completion}). The final networks
after training exhibit the expected pattern: each input node sends one strong,
fixed connection to the corresponding output node, as well as one plastic
connection to every output node. As a result, during pattern presentation, each
non-zero input develops a strong connection to every non-zero output due to
Hebbian learning, ensuring successful pattern completion on the second step when
one of the nonzero inputs is stimulated. 

\subsection{One-shot learning of arbitrary patterns}

In this task, at each episode, the network must learn to associate each of two
random binary vectors with its label. The labels are simply two-element vectors
set to 01 for one of the
vectors, and 10 for the other. Importantly, learning is one-shot: at the first
timestep, the input consists of the first pattern, suffixed with label 01;
and at the second timestep, the input vector is the second pattern, suffixed
with label 10. These are the only times the labels are presented as inputs: at
all other times, the input is one of the patterns, suffixed with the neutral
suffix 00, and the network's output must be the label associated with the current
pattern.

\begin{figure}[b]
\noindent
\begin{subfigure}[t]{0.35\textwidth}
\noindent
\includegraphics[scale=0.5]{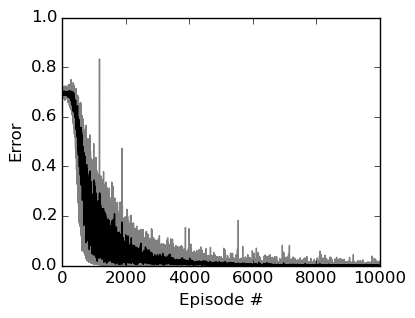}
\subcaption{Training error}
\end{subfigure}
\begin{subfigure}[t]{0.4\textwidth}
\centering
\begin{tikzpicture}[scale=.9]
\centering
\begin{scope}[every node/.style={circle,thick,draw}]
\node (Z1) at (1,3) {\tiny Z1};
\node (Z2) at (3,3) {\tiny Z2};
\node (Y1) at (1,1.5) {\tiny Y1};
\node (Y2) at (3,1.5) {\tiny Y2};
\node (L1) at (0,0) {\tiny L1};
\node (L2) at (1,0) {\tiny L2};
\node (X1) at (2.5,0) {\tiny X1};
\node (X2) at (3.5,0) {\tiny X2};
\node (X3) at (4.5,0) {\tiny X3};
\end{scope}
\begin{scope}[
        every node/.style={fill=white,circle},
    every edge/.style={draw=black,very thick}]
    
    
    \path [->] (L1) edge [blue,opacity=0.33,bend left=10]  (Y1);
    \path [->] (L1) edge  [bend left=10] (Y2);

    \path [->] (L2) edge [blue,opacity=0.33,bend left=10]  (Y2);
    \path [->] (L2) edge [bend left=10]  (Y1);

    \path [dashed,thin][->] (X1) edge   (Y1);
    \path [dashed,thin][->] (X2) edge   (Y1);
    \path [dashed,thin][->] (X3) edge   (Y1);

    \path [dashed,thin][->] (X1) edge   (Y2);
    \path [dashed,thin][->] (X2) edge   (Y2);
    \path [dashed,thin][->] (X3) edge   (Y2);

    \path [->] (Y1) edge  (Z2);
    \path [->] (Y1) edge [blue,opacity=0.33] (Z1);
    \path [->] (Y2) edge  (Z1);
    \path [->] (Y2) edge [blue,opacity=0.33] (Z2);

\end{scope}
\end{tikzpicture}
\subcaption{Typical trained network}
\end{subfigure}
\begin{minipage}[b]{0.2\textwidth}
\noindent
\small
\begin{itemize}[leftmargin=*] 
\item[]\tikz{\path[very thick,->,draw=black] 
        (0,0) -- (1,0) ;}Excitatory fixed-weight connection
\item[]\tikz{\path[very thick,->,draw=blue,opacity=0.33] 
        (0,0) -- (1,0) ;}Inhibitory fixed-weight connection
\item[]\tikz{\path[very thick,->,dashed,draw=black] 
        (0,0) -- (1,0) ;}Excitatory plastic connection
\item[]\tikz{\path[very thick,->,dashed,draw=blue,opacity=0.33] 
        (0,0) -- (1,0) ;}Inhibitory plastic connection
\end{itemize}
\end{minipage}

\caption{Results for the one-shot learning experiment. (a) Median absolute error
    per timestep over each episode. Conventions are as in Figure
    \ref{fig:completion}.
(b) Schema of a
typical network after training. In addition to the label nodes L1 and L2, only 3 pattern elements shown for clarity (actual
pattern size: 8 elements). See text for details.}
\label{fig:oneshot}
\end{figure}
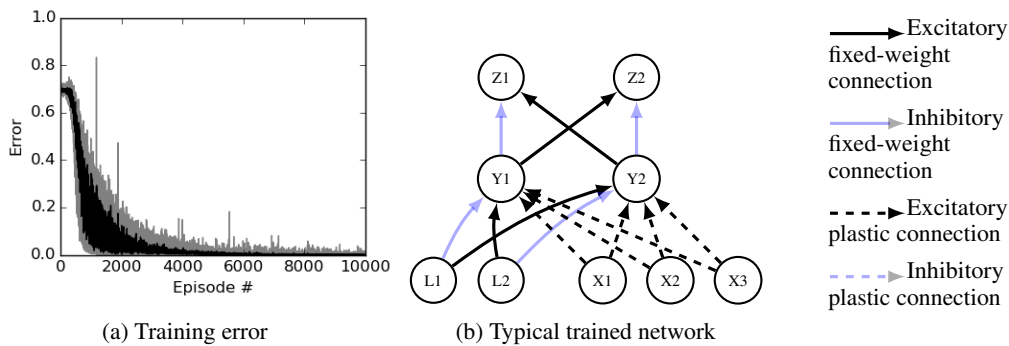

Patterns are random vectors of $N$ elements, each having
value 1 or -1, with at least one position differing between the two patterns to
be learned ($N=8$ for all experiments). The networks have an input layer ($N+2$ nodes), a hidden layer (2 nodes), and an
output layer (2 nodes). For simplicity, only the first layer of weights
(input-to-hidden) can have plasticity. The final layer implements softmax
competition between the nodes. Each episode lasts 20 timestep, of which only the first
two contain the expected label for each pattern. We use cross-entropy loss
between the output values and the expected label at each time step, except for
the 2-step learning period during which network output is ignored.

Again, the algorithm reliably learns to solve the task (Figure
\ref{fig:oneshot}). The trained networks are
organized in such a way that one hidden node learns the pattern with label 01,
and the other learns the pattern associated with label 10: they receive strong,
fixed (positive and negative) connections from the label bits, but receive only
strong plastic connections (with zero fixed-weight connections) from the pattern
bit. This allows each node to be imprinted to the corresponding pattern. The weights between hidden and top layer ensures that the top two nodes
produce the adequate label.

Note that some networks displayed a somewhat different pattern where all
connections between pattern nodes and hidden nodes have negative plasticity
coefficients. We discuss this configuration in the next subsection.

\subsection{Reversal learning}

Previous experiments show that the algorithm can train networks to learn fast
associations of environmental inputs. But can it also teach networks to adapt to
a changing environment - that is, to perform continual learning over their
lifetime? 

To test this, we adapt the previous one-shot learning task into a
continual learning task: halfway through each episode, we invert the two
patterns, so that the pattern previously associated with label 01 is now
associated with label 10, and vice-versa. We show each of the pattern with its
updated label once. Then we resume showing input patterns with neutral, 00
suffixes, and expect the network's output to be the adequate new label for
each input pattern.

\begin{figure}
\noindent
\begin{subfigure}[t]{0.35\textwidth}
\noindent
\includegraphics[scale=0.5]{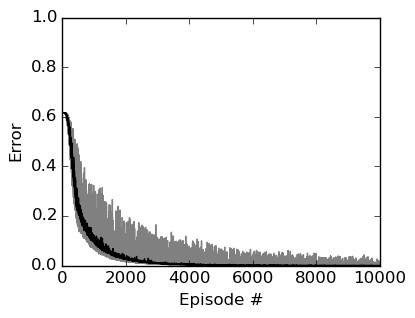}
\subcaption{Training error}
\end{subfigure}
\begin{subfigure}[t]{0.4\textwidth}
\centering
\begin{tikzpicture}[scale=.9]
\centering
\begin{scope}[every node/.style={circle,thick,draw}]
\node (Z1) at (1,3) {\tiny Z1};
\node (Z2) at (3,3) {\tiny Z2};
\node (Y1) at (1,1.5) {\tiny Y1};
\node (Y2) at (3,1.5) {\tiny Y2};
\node (L1) at (0,0) {\tiny L1};
\node (L2) at (1,0) {\tiny L2};
\node (X1) at (2.5,0) {\tiny X1};
\node (X2) at (3.5,0) {\tiny X2};
\node (X3) at (4.5,0) {\tiny X3};
\end{scope}
\begin{scope}[
        every node/.style={fill=white,circle},
    every edge/.style={draw=black,very thick}]
    
    
    \path [->] (L1) edge [blue,opacity=0.33,bend left=10]  (Y2);
    \path [->] (L1) edge  [bend left=10] (Y1);

    \path [->] (L2) edge [blue,opacity=0.33,bend left=10]  (Y1);
    \path [->] (L2) edge [bend left=10]  (Y2);

    \path [dashed,thin][->] (X1) edge  [blue,opacity=0.33] (Y1);
    \path [dashed,thin][->] (X2) edge  [blue,opacity=0.33] (Y1);
    \path [dashed,thin][->] (X3) edge  [blue,opacity=0.33] (Y1);

    \path [dashed,thin][->] (X1) edge   [blue,opacity=0.33](Y2);
    \path [dashed,thin][->] (X2) edge   [blue,opacity=0.33](Y2);
    \path [dashed,thin][->] (X3) edge   [blue,opacity=0.33](Y2);

    \path [->] (Y1) edge  (Z2);
    \path [->] (Y1) edge [blue,opacity=0.33] (Z1);
    \path [->] (Y2) edge  (Z1);
    \path [->] (Y2) edge [blue,opacity=0.33] (Z2);

\end{scope}
\end{tikzpicture}
\subcaption{Typical trained network}
\end{subfigure}
\begin{minipage}[b]{0.2\textwidth}
\noindent
\small
\begin{itemize}[leftmargin=*] 
\item[]\tikz{\path[very thick,->,draw=black] 
        (0,0) -- (1,0) ;}Excitatory fixed-weight connection
\item[]\tikz{\path[very thick,->,draw=blue,opacity=0.33] 
        (0,0) -- (1,0) ;}Inhibitory fixed-weight connection
\item[]\tikz{\path[very thick,->,dashed,draw=black] 
        (0,0) -- (1,0) ;}Excitatory plastic connection
\item[]\tikz{\path[very thick,->,dashed,draw=blue,opacity=0.33] 
        (0,0) -- (1,0) ;}Inhibitory plastic connection
\end{itemize}
\end{minipage}

\caption{Results for the reversal learning experiment. Conventions are as in Figure
    \ref{fig:completion}.
(a) Mean absolute error
    per timestep over each episode. (b) Schema of a
typical network after training. Notice the negative plasticity connections from
the pattern nodes to the hidden layer. See text for details.}
\label{fig:reversal}
\end{figure}
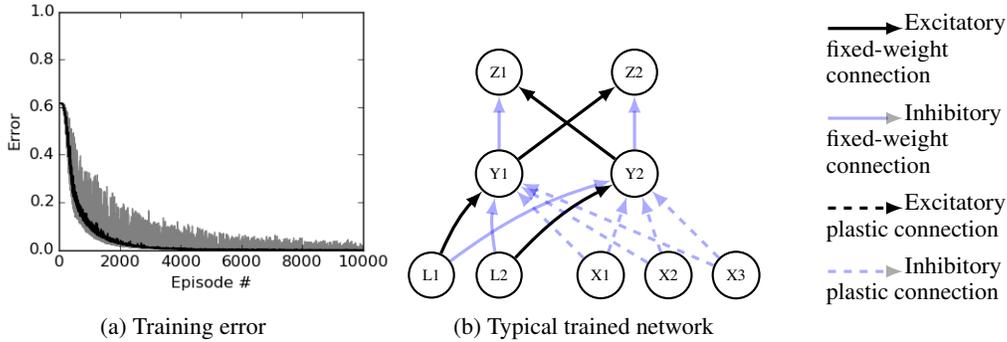

The algorithm also successfully learns to solve this problem (Figure
\ref{fig:reversal}). The networks are somewhat similar to the ones obtained in
one-shot
learning, but with an important difference: connections from pattern input
to hidden nodes now consistently have \emph{negative} plasticity coefficients. While some
networks trained for the one-shot learning task also had this feature,
all networks trained  for reversal learning consistently show it. This seems to
be a crucial feature for reversal learning, because clipping plasticity coefficients to positive values
prevents learning in this task while still allowing successful learning in the
one-shot task (data not shown). What does reversal learning seem to require
negative plasticity? 

Negative plasticity implies that hidden layer firing will be anti-correlated to
the presence of the imprinted stimulus. This in itself is unlikely to have any
effect, since the effect can be negated by switching signs in the output layer or label node inputs.
However, negative plasticity also has the important consequence of making the
Hebbian traces \emph{decrease} over time, rather than increase as they would if
plasticity coefficient were positive. 

It is well-known that Hebbian plasticity
creates a positive feedback: correlation between input and output increases the
connection weight, which in turn increases the correlation in firing, etc. This would
pose a problem for one-shot reversal learning, because by the time the new patterns are
shown, the existing associations would be too strong to be erased in a single
timestep.
However, with negative plasticity coefficients, the opposite is true: the large
Hebbian trace created on initial imprinting becomes self-decreasing due to a
\emph{negative} feedback loop. As a result, the Hebbian traces created by the first
association decrease over time. This has little effect on ongoing responses, since
the output nonlinearities will magnify even small differences to
produce the correct responses; for the same reason, one-shot learning (in which
there is no reversal) is mostly indifferent to the sign of plasticity coefficients,
since the episodes are short enough that the final Hebbian traces will always be
large enough to support correct choice. However, in the case of reversal
learning, decreasing Hebbian traces are vital: when the second association is shown,
the existing Hebbian traces  are now small enough to be completely erased (and
indeed reversed) in a single presentation, which would not be the case under
positive plasticity and increasing Hebbian traces.

In short, the BOHP method has not only determined which connections must be
plastic to learn an association; it can also develop a precise fine-tuning of
the temporal dynamics of this plasticity, by modulating the sign of plasticity
coefficients. This remarkable result confirms the potential of BOHP to deal with
temporal dynamics in environmental learning.

\section{Conclusions and future work}

In this paper we have introduced a method for designing networks endowed with
Hebbian plasticity through gradient descent and backpropagation. This method
allows the network to ``learn how to learn'' in order to solve a problem with
unpredictable parameters. The method successfully solved simple learning tasks, including
one-shot and reversal learning.

In this expository we only use a very simple form of Hebbian plasticity, namely,
the running average of the product between pre- and post-synaptic activities.
However, there are other possible formulations of Hebbian plasticity, such as
covariance learning (mean-centering pre- and post-synaptic responses), instar
and outstar rules, or BCM learning. These can be implemented in BOHP by
rewriting the gradient equations appropriately, which might expand the
capacities of BOHP. However, as shown above, the simple Hebbian plasticity used
here can already produce fast, reliable lifelong learning.

Furthermore, the networks shown here use fixed plasticity constants to determine
the strength of plasticity. However, in biological brains, plasticity is
modulated over time by various neuromodulators (especially dopamine,
acetylcholine and norepinephrine), which are themselves under neural control.
Thus, biological brains can decide not just where, but \emph{when} to apply
plasticity, which is crucial for learning complex behaviors. While
neuromodulation has been used in neural networks built by evolutionary methods
\cite{Soltoggio2013-rg}, the methods described in this paper could be extended to allow
backpropagation to design modulable-plasticity networks.


In conclusion, we suggest that backpropagation of Hebbian plasticity is an
efficient way to endow neural networks with long-term memories and lifelong
learning abilities. 

\section*{Appendix}

Here we provide a derivation of the gradients of output cell
response $y$ at a given timestep $t_z$ with regard to the $\alpha$ coefficient
of an incoming connection $k$ (where input activity of the pre-synaptic neuron
at this connection is denoted by $x_k$). 

First we simply write out the
full expression for $y$, from Equation \ref{eq:y} (again, we initially omit
the $\tanh$ nonlinearity):

\[
\frac{\partial y(t_z)}{\partial \alpha_k} = \frac{\partial }{\partial \alpha_k}[ \sum_{l \in inputs}w_l x_l(t_z) + \sum_{l \in inputs}\alpha_l Hebb_l(t_z) x_l(t_z) ] 
\]

The first summand on the right-hand side denotes the inputs to $y$ from incoming connections through the fixed weights; since this term does not depend on $\alpha$ in any way, we can remove it from the gradient computation.

The second summand denotes inputs through plastic connections. The cases for
$l=k$ and $l\neq k$ must be handled differently, since we are differentiating
with regard to $\alpha_k$:

\begin{align}
\frac{\partial y(t_z)}{\partial \alpha_k} &= \frac{\partial }{\partial \alpha_k}[
\sum_{l \neq k }\alpha_l Hebb_l(t_z) x_l(t_z) + \alpha_k Hebb_k(t_z) x_k(t_z)]\\
&= \sum_{l \neq k }[\frac{\partial }{\partial \alpha_k} (\alpha_l x_l(t_z) Hebb_l(t_z))] + \frac{\partial }{\partial \alpha_k}[\alpha_k Hebb_k(t_z) x_k(t_z)]
\end{align}

With regard to $\alpha_k$, the derivative in the first right-hand-side term has the
form $d(Const*f(\alpha_k))/d\alpha_k$, since only the $Hebb_l(t_z)$ depends on $\alpha_k$
(indirectly through $y$), while neither $\alpha_l$ nor $x_l(t_z)$ does. By contrast, the second right-hand-side term has the
form $d(Const*\alpha_k*f(\alpha_k))/d\alpha_k$, so we must develop it using the identity
$(xf(x))'=xf'(x)+f(x)$. Therefore:

\[
\frac{\partial y(t_z)}{\partial \alpha_k} = \sum_{l \neq k }[\alpha_l
x_l(t_z)\frac{\partial }{\partial \alpha_k}Hebb_l(t_z)] +
x_k(t_z)(\alpha_k\frac{\partial }{\partial \alpha_k}Hebb_k(t_z) + Hebb_k(t_z))
\]

Replacing the $Hebb(t)$ terms by their full expression as the accumulated product of $x$ and
$y$ (Eq. \ref{eq:hebb}), we get:

\begin{align}
\frac{\partial y(t_z)}{\partial \alpha_k} &= \sum_{l \neq k }[\alpha_l x_l(t_z) \frac{\partial }{\partial
\alpha_k}\sum_{t_u < t_z}(1-\gamma)\gamma^{tz-tu}x_l(t_u)y(t_u))] \nonumber \\ 
& \qquad {} + x_k(t_z)[\alpha_k\frac{\partial}{\partial \alpha_k}\sum_{t_u <
t_z}(1-\gamma)\gamma^{tz-tu}x_k(t_u)y(t_u) + Hebb_k(tz)]\\
&= \sum_{l \neq k }[\alpha_l x_l(t_z) \sum_{t_u < t_z}(1-\gamma)\gamma^{tz-tu}x_l(t_u)\frac{\partial }{\partial
\alpha_k}y(t_u)] \nonumber \\
& \qquad {} + x_k(t_z)[\alpha_k\sum_{t_u <
t_z}x_k(t_u)(1-\gamma)\gamma^{tz-tu}\frac{\partial}{\partial \alpha_k}y(t_u) + Hebb_k(tz)]\\
&= \sum_{l \in inputs}[\alpha_l x_l(t_z)\sum_{t_u <
t_z}(1-\gamma)\gamma^{t_z-t_u}x_l(t_u)\frac{\partial
}{\partial
\alpha_k}y(t_u)] +
x_k(t_z)Hebb_k(tz)
\end{align}

where in the last equation above, $l$ runs over all incoming connections to $y$,
including $k$. This recursive gradient equation is identical to Eq.
\ref{eq:gradalpha}.

Eq. \ref{eq:gradw} is derived much in the same manner (though slightly simpler
since we do not need to use the $(xf(x))'=xf'(x)+f(x)$ identity). For future applications to
many-layers networks, equations for the gradient
$\frac{\partial y(t_z)}{\partial x_k}$ are easily obtained with a similar
derivation.

\small
\printbibliography

\end{document}